\documentclass{article}

\usepackage[final]{corl_2023} % Uncomment for the camera-ready ``final'' version.
\usepackage{amsmath,amssymb}
\usepackage{adjustbox}
\usepackage{booktabs,graphicx,makecell,siunitx,eqparbox}
\usepackage{tabularx}
\usepackage[table]{xcolor}  
\usepackage{subfig}
\usepackage{graphicx}
\usepackage{wrapfig}
\usepackage[capitalize,noabbrev]{cleveref}

\usepackage[belowskip=-20pt,aboveskip=0pt]{caption}

\usepackage[nopostdot,nogroupskip,nonumberlist]{glossaries}
\makeglossaries
\newglossaryentry{auc}{name=AUC, description={Area Under Curve},first={Area Under Curve (AUC)}}
\newglossaryentry{auroc}{name=AUROC, description={Area Under Receiver Operation Curve},first={Area Under Receiver Operation Curve (AUROC)}}
\newglossaryentry{bnn}{name=BNNs, description={Bayesian Neural Networks},first={Bayesian Neural Networks (BNNs)}}
\newglossaryentry{ci}{name=CI, description={Mutual Cross-Information},first={Mutual Cross-Information (CI)}}
\newglossaryentry{crsb}{name=cRSB, description={Conditional Resampled Base Distributions},first={Conditional Resampled Base Distributions (cRSB)}}
\newglossaryentry{dl}{name=DL, description={Deep Learning},first={Deep Learning (DL)}}
\newglossaryentry{em}{name=EM, description={Expectation Maximization},first={Expectation Maximization (EM)}}
\newglossaryentry{fpr}{name=FPR, description={False Positive Rate},first={False Positive Rate (FPR)}}
\newglossaryentry{gm}{name=MoG, description={Mixture of Gaussians Distributions},first={Mixture of Gaussians (MoG)}}
\newglossaryentry{gmm}{name=GMMs, description={Gaussian Mixture Models},first={Gaussian Mixture Models (GMMs)}}
\newglossaryentry{ib}{name=IB, description={Information Bottleneck},first={Information Bottleneck (IB)}}
\newglossaryentry{id}{name=ID, description={In-Distribution},first={In-Distribution (ID)}}
\newglossaryentry{kld}{name=KLD, description={Kullaback-Leibler Divergence},first={Kullaback-Leibler Divergence (KLD)}}
\newglossaryentry{ll}{name=LL, description={Log-Likelihood},first={Log-Likelihood (LL)}}
\newglossaryentry{lars}{name=LARS, description={Learned accept/reject sampling},first={Learned accept/reject sampling (LARS)}}
\newglossaryentry{mle}{name=MLE, description={Maximum Likelihood Estimation},first={Maximum Likelihood Estimation (MLE)}}
\newglossaryentry{mi}{name=MI, description={Mutual Information},first={Mutual Information (MI)}}
\newglossaryentry{ml}{name=ML, description={Machine Learning},first={Machine Learning (ML)}}
\newglossaryentry{nf}{name=NFs, description={Normalizing Flows},first={Normalizing Flows (NFs)}}
\newglossaryentry{ood}{name=OOD, description={Out-of-Distribution},first={Out-of-Distribution (OOD)}}
\newglossaryentry{rsb}{name=RSB, description={Resampled Base Distributions},first={Resampled Base Distributions (RSB)}}
\newglossaryentry{tpr}{name=TPR, description={True Positive Rate},first={True Positive Rate(TPR)}}

\def\baseParam{\psi}
\def\baseVarCap{Z}
\def\baseVar{\bb{z}}
\def\flowFunc{T}

\def\flowParam{\phi}

\def\featDim{d} 
\def\featVarCap{U}
\def\featVar{\bb{u}}
\def\detFunc{F} 
\def\detParam{\theta}

\newcommand{\rebuttal}[1]{\textcolor{black}{#1}}
\newcommand{\bb}[1]{\textbf{#1}}
\newcommand{\pb}[1]{\textbf{#1}^\prime}
\newcommand{\mb}[1]{\mathcal{#1}}

\usepackage{booktabs}
\newcommand{\ra}[1]{\renewcommand{\arraystretch}{#1}}

\newcommand\myeq{\mkern1.5mu{=}\mkern1.5mu}

\usepackage{array}
\newcolumntype{L}[1]{>{\raggedright\let\newline\\\arraybackslash\hspace{0pt}}m{#1}}
\newcolumntype{C}[1]{>{\centering\let\newline\\\arraybackslash\hspace{0pt}}m{#1}}
\newcolumntype{R}[1]{>{\raggedleft\let\newline\\\arraybackslash\hspace{0pt}}m{#1}}

\usepackage{setspace}
\title{Topology-Matching Normalizing Flows for Out-of-Distribution Detection in Robot Learning}
\author{
  Jianxiang Feng$^{*, 1}$
  Jongseok Lee$^{2,3}$, 
  Simon Geisler$^1$,
  Stephan Günnemann$^1$,
  Rudolph Triebel$^{2,3}$
  \\
  $^1$ Department of Informatics, Technical University of Munich (TUM)
  \\
  $^2$Institute of Robotics and Mechatronics, German Aerospace Center (DLR)  
  \\
  $^3$ Department of Informatics, Karlsruhe Institute of Technology (KIT)
  \\
  \texttt{jianxiang.feng@tum.de, \{jongseok.lee, rudolph.triebel\}@dlr.de},
  \\
  \texttt{\{geisler, guennemann\}@in.tum.de}\\
}

\begin{document}
\maketitle
%===============================================================================

% Reliably deploying an object detector on robots in the wild necessitates the need of accurate and fast \gls{ood} identification. 
\begin{abstract}
    To facilitate reliable deployments of autonomous robots in the real world, Out-of-Distribution (OOD) detection capabilities are often required. 
    A powerful approach for OOD detection is based on density estimation with Normalizing Flows (NFs).
    % To facilitate reliable deployments of autonomous robots in the wild, Normalizing Flows (NFs) is powerful density-based approach to equip the often required Out-of-Distribution (OOD) detection capabilities.
    However, we find that prior work with NFs attempts to match the complex target distribution topologically with na\"ive base distributions leading to adverse implications. 
    In this work, we circumvent this topological mismatch using an expressive class-conditional base distribution trained with an information-theoretic objective to match the required topology. % incorporate task-specific details e.g.\ class conditions. 
    % Moreover, we train it with an information-theoretic objective to effectively incorporate task-specific details e.g.\ class conditions, commonly ignored in pure generative modeling. 
    %,to match the required topology.
    %In this work, we circumvent this topological mismatch using expressive class-conditional base distributions that can learn to match the required topology. % , including a learnable number of modes. 
    %%We train the more flexible base distribution with the Information Bottleneck objective. 
    The proposed method enjoys the merits of wide compatibility with existing learned models without any performance degradation and minimum computation overhead while enhancing OOD detection capabilities. %efficient runtime inference and cost-effective memory overheads. 
    We demonstrate superior results in density estimation and 2D object detection benchmarks in comparison with extensive baselines. 
    Moreover, we showcase the applicability of the method with a real-robot deployment. 
    %We empirically demonstrate the benefits in synthetic density estimation and 2D object detection benchmarks as well as a real-robot deployment. 

    %To facilitate reliable deployment of object detectors in the wild, Normalizing Flows (NFs) is promising to gracefully handling Out-of-Distribution (OOD) objects by detecting them beforehand. 
    %However, its modeling ability is limited by the topological mismatch problem, i.e. the commonly complex target distribution should topologically match the often simple base distribution. 
    %Meanwhile traditionally studied OOD approaches with NFs do not seem ideal for class-structured predictions in object detection. 
    %In this work, we propose a method for fast and effective OOD-aware object detection with topologically matched NFs. 
    %Specifically, we introduce an impressive base distribution that is capable of mitigating the topological restriction. 
    %To better take into account class information in object detection, we train the more flexible base distribution with the Information Bottleneck objective. 
    %The proposed method enjoys the merits of general compatibility with existing object detectors, efficient run-time inference and cost-effective memory consumption. 
    %Experimentally, we provide empirical evidences to show the benefits and practicality through comprehensive evaluation on density estimation, object detection and real-robot  deployment. 
    %
    %
\end{abstract}

\let\thefootnote\relax\footnotetext{$^*$: work done when working at DLR.}
\let\thefootnote\relax\footnotetext{code: \href{https://github.com/DLR-RM}{https://github.com/DLR-RM}}

%===============================================================================
\section{Introduction}
\label{intro}
% With the widespread adoption of \gls{dl} models in modern robotic software stacks, coping with \gls{ood} data that is not well represented in the training set poses a pressing challenge on the path towards trustworthy open-world autonomy. 
% Such models may behave unreliably when confronted with unknowns in uncontrolled and unpredictable outside-the-lab environments~\cite{yang2021generalized}, where our robots such as self-driving cars~\cite{nitsch2021out}, delivery drones~\cite{lee2022trust} or healthcare robots \cite{feng2022bayesian} are envisioned to perform complex tasks. 
% Robustly identifying \gls{ood} data that is not well represented in the training set poses a pressing challenge on the path towards trustworthy open-world robotic systems such as self-driving cars~\cite{nitsch2021out}, delivery drones~\cite{lee2022trust} or healthcare robots \cite{feng2022bayesian}.
The reliable identification of \gls{ood} data, which is not well represented in the training set, poses a pressing challenge on the path towards trustworthy open-world robotic systems such as self-driving cars~\cite{nitsch2021out}, delivery drones~\cite{lee2022trust} or healthcare robots \cite{feng2022bayesian}.
For example, with widespread adoption in the perception pipeline, existing object detectors have been reported to over-confidently misclassify an \gls{ood} object into a known class, which might obfuscate the decision-making module and eventually cause catastrophic consequences in safety-critical scenarios~\cite{nitsch2021out, Dhamija_2020_WACV, sinha2022system}. 

%For \gls{ood} detection, \gls{nf} are a popular class of generative models~\cite{kirichenko2020normalizing, zhang2020hybrid, li2022out, mackowiak2021generative}. \gls{nf} may represent complex probability distribution~\cite{papamakarios2021normalizing} by learning a series of transformations from a simple base distribution to a complex target distribution.
\gls{nf} are a popular class of generative models~\cite{kirichenko2020normalizing, zhang2020hybrid, li2022out, mackowiak2021generative} that may be used for \gls{ood} detection. 
\gls{nf} represent complex probability distributions~\cite{papamakarios2021normalizing} with a learnable series of transformations from a simple base distribution to a complex target distribution.
However, \gls{nf}' expressivity~\cite{cornish2020relaxing, wu2020stochastic, stimper2022resampling} and numerical stability~\cite{behrmann2021understanding, hagemann2021stabilizing} is limited by a fundamental constraint: the supports of the base and target distribution should preserve \textit{similar topological properties} (Definition 3.3.10 in~\citet{Runde05}). 
The topological properties subsume different geometrical characteristics of the target distribution, including its continuity, the number of connected components, or the number of modes.
%The topological properties subsume different geometrical characteristics of the target distribution, including its continuity, the number of connected components or "holes", or how they are "knotted".
%A topology describes how elements in a set (e.g., samples drawn from a distribution) relate to each other with properties like connectedness, continutiy to name a few. 
% This complicates the modeling of \gls{nf} for complex distribution in tasks like object detection. % where the topological properties are unknown and hard to reason about.
Increasing the capacity of the transformation may mitigate this constraint. Yet, this raises computation and memory demands~\cite{cornish2020relaxing, jaini2020tails, wu2020stochastic}. 
An alternative to overcome the topological mismatch is to increase the flexibility of the base distribution, which is surprisingly under-explored in the \gls{ood} detection literature. 

Therefore, we propose to equip \gls{nf} with efficient but flexible base distributions for \gls{ood} detection in robot learning.
%In this work, we propose to achieve this by utilizing topologically matched \gls{nf}. 
\begin{figure}[!ht]
    \centering
    \vspace{43pt}
    \includegraphics[width=1.0\linewidth]{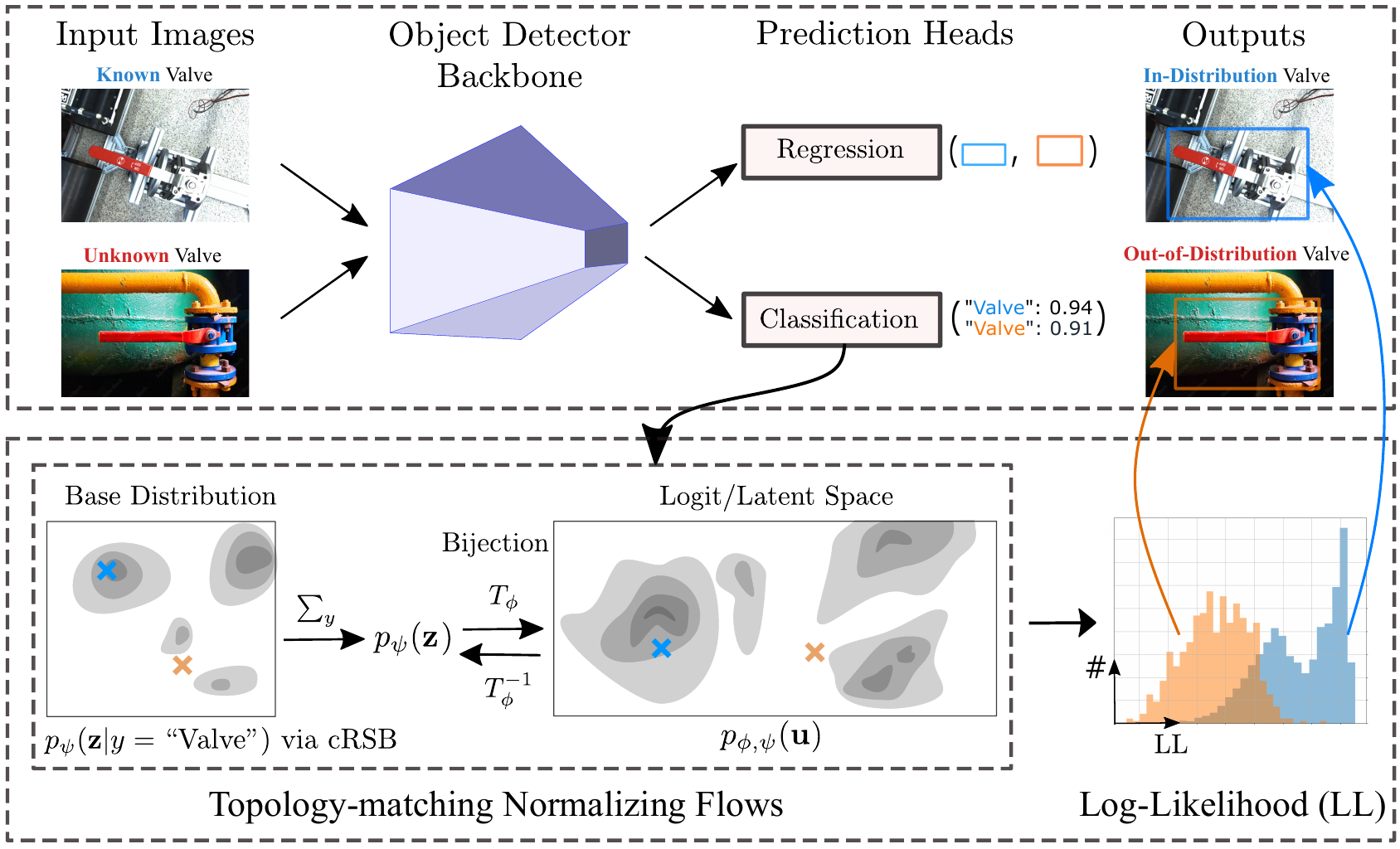}
    \caption{\textbf{The proposed architecture}. We overcome the topological mismatch problem in \gls{nf} to accurately model \gls{id} density. That is, the \gls{crsb} base distribution trained with \gls{ib} \(p_{\psi}(\mathbf{z|y})\) can, e.g., adapt the numbers of modes to match target distribution with complex topology. 
    Then we can identify \gls{ood} objects by low predicted log-likelihoods more reliably (best viewed in color). %Therefore, \gls{ood} objects can be identified by low predicted log-likelihoods to prevent potential disastrous consequences in safety-critical robotic applications, e.g. intricate manipulation of a known valve (best viewed in color).
    }
    \label{fig:method_fig}
\end{figure}
Concretely, we replace the frequently used uni-modal Gaussian base distribution with the \gls{crsb}, a class-conditional version of a learnable base distribution for mitigating the topological problem in \gls{nf} -- \gls{rsb}~\cite{stimper2022resampling}. 
\gls{crsb} can learn the required topological properties, like adapting the number of modes, to match the unknown topological structure of the latent class-specific target distribution (\cref{fig:method_fig}).
Moreover, %in contrast to class-conditional \gls{gm} used in previous work, 
we adapt our \gls{crsb} with an adapted \gls{ib} objective~\cite{ardizzone2020training} to balance fusing class-conditional information with the marginalized density estimation capabilities in \gls{nf}. \gls{ib}~\cite{tishby2000information} is an information-theoretic objective to incorporate task-specific details e.g. class conditions, which are commonly ignored in pure generative modeling. This delivers a topology in the base distribution that is more accurately aligned to the one in the target distribution (see \cref{fig:density_toy}).
%We achieve this by %learning an expressive base distributions 
%a learnable base distribution and incorporating class-structured information with \gls{ib} as training objective. 
%Concretely, to address the topological problem of \gls{nf}, we replace the frequently used uni-modal Gaussian base distribution with \gls{rsb}~\cite{stimper2022resampling}. 
%To utilize the class information inherent in object detectors, we further devise \gls{crsb} to innately enable conditional modeling and further adapt with the formulation of \gls{ib} optimization objective for \gls{ood} detection.

Our \gls{ood} detection approach using topology-matching \gls{nf} is powerful and yet resource-efficient for open-set object detection. It is applicable to diverse object detectors (e.g., Faster-RCNN~\cite{ren2015faster} and Yolov7~\cite{wang2023yolov7} used in this work) with minor changes and no loss of prediction performance. 
%Our resulting topologically matched NF has %\gls{nf} have 
%several desirable features for object detection in open-set conditions. First, our \gls{ood} detection via \gls{nf} does not require changing the %First, our \gls{nf} do not incur changes to the existing network architecture and therefore, we can plug it into already learned \gls{dl} models for \gls{ood} detection. 
Moreover, our approach is sampling-free, i.e., only a single forward pass is required for efficient test-time inference while keeping the space memory tractable. 
As a result, our method is suitable for robotic applications that require a fast and robust perception module.
We empirically show the state-of-the-art performance of the proposed idea using synthetic density estimation and 2D object detection tasks against extensive baselines. %based on two adapted public benchmark datasets (Pascal-VOC~\cite{pascalVOC} and COCO~\cite{lin2014microsoft}). 
To further validate the applicability in robotics, we examine an object detector equipped with the proposed method on an exemplary inspection and maintenance aerial robot, showing the practical benefits of negligible memory and run-time overhead. 

\textbf{Contributions.} Our main contribution is a \gls{nf}-based \gls{ood} detection method that overcomes the topological constraints while taking class-conditional information into account. 
We show that training with \gls{ib} yields effective representation with superior \gls{ood} detection capabilities.
%To summarize, our main contribution is to propose an approach for \gls{nf}-based \gls{ood} detection by overcoming the fundamental topological constraints and taking into account the class conditional information with \gls{ib}. 
%In experiments, we 
We conduct a comprehensive empirical evaluation using both synthetic density estimation and public object detection datasets followed by a real-world robot deployment, which overall shows the effectiveness of the proposed approach.

\section{Methodology}
\label{methods}
%Our method for \gls{ood} detection in the demanding realm of Open-Set Object Detection is well-suited for mobile applications due to its resource efficiency. 
%Before we introduce our main method and describe how it overcomes the topological mismatch, in \Cref{tp_problem}, we will first formalize the setting. Last, in \Cref{inference}, we describe specifics about how the resulting Normalizing Flow is used for \gls{ood} detection.

% We define a data-set $\mathcal{D}_{id}=\{ (\bb{x}_i, \bb{b}_i, y_i) \}_{i=1}^{N}$ consisting of $N$ samples of objects with corresponding object bounding box coordinates $\bb{b}_i \in \mb{R}^4$ and class label $y_i \in \mb{Y} = \{1,2,...,C\}$ on an image $\bb{x}_i \in \mb{X}_{id}$ drawn from some known distribution $\mb{P}_{id}=\mb{X}_{id} \times \mb{Y} \times \mb{R}^4$. 

\paragraph{Problem Formulation} 
Given an image $\bb{x} \in \mb{X}$ and a trained object detector \(\detFunc_\detParam\) that localizes a set of objects with corresponding bounding box coordinates $\bb{b}_i \in \mb{R}^4$ as well as class label $y_i \in \mb{Y} = \{1,2,...,C\}$, the task is to distinguish if $(\bb{x}, \bb{b}_i, y_i)$ is \gls{id}, i.e., drawn from $\mb{P}_{id}$, or \gls{ood}, i.e., belongs to the unknown distribution $\mb{P}_{ood}$. 
For conciseness, from now on we omit the suffix $i$ and use $y$ to denote the class label without further notice.
As discussed, a powerful \gls{ood} detection can be obtained via density estimation using \gls{nf}. This density estimator identifies \gls{ood} objects with low likelihoods after being trained \textit{only} on data drawn from $\mb{P}_{id}$. Following relevant prior~\citep{wei2022logitnorm, miller2021uncertainty}, we use the semantically rich logit space (pre-softmax layer) for density estimation. To note that, our method can be readily applied to other (high-dimensional) latent feature spaces.

\begin{wrapfigure}[12]{r}{0.55\textwidth}
   \centering
   \vspace{-23pt}
   \subfloat[\(p(\featVar|y=0)\)\label{fig:filaments:a}]{\includegraphics[width=0.33\linewidth]{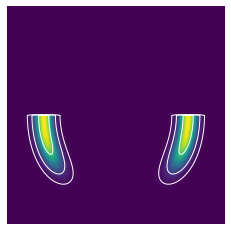}}
   \hfill
   \subfloat[\(p_{\flowParam, \baseParam}(\featVar|y\myeq 0)\)\label{fig:filaments:b}]{\includegraphics[width=0.33\linewidth]{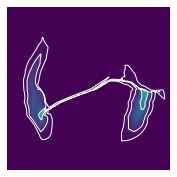}}
   \hfill
   \subfloat[\(p_{\baseParam}(\baseVar|y=0)\)\label{fig:filaments:c}]{\includegraphics[width=0.33\linewidth]{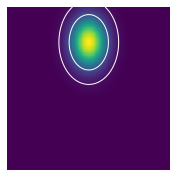}}
   %
   % \\
   % %
   % \subfloat[\(p(\featVar|y=0)\)\label{fig:filaments:a}]{\includegraphics[width=0.33\linewidth]{materials/toy/moons_gt_density_0.png}}
   % \hfill
   % \subfloat[\(p_{\flowParam, \baseParam}(\featVar|y=0)\)\label{fig:filaments:b}]{\includegraphics[width=0.33\linewidth]{materials/toy/moons_resampled_cls_out_wib_density_0.png}}
   % \hfill
   % \subfloat[\(p_{\baseParam}(\baseVar|y=0)\)\label{fig:filaments:c}]{\includegraphics[width=0.33\linewidth]{materials/toy/moons_resampled_cls_out_wib_base_0.png}}
   \caption{Filament connect modes in the modeled class-conditional distribution (b) if using (trainable) uni-modal base (c) for the multi-modal target (a).}
   \label{fig:filaments}
\end{wrapfigure}
\gls{nf} are known to be universal distribution approximators~\cite{papamakarios2021normalizing}. 
That is, they can model a complex target distribution $p(\featVar)$ on a space $\mb{R}^\featDim$ by defining $\featVar$ as a transformation $\flowFunc_\flowParam: \mb{R}^\featDim \rightarrow \mb{R}^\featDim$ from a well-defined base distribution $p_{\baseParam}(\baseVar)$, where $\flowParam$ and $\baseParam$ are model parameters, respectively: 
\begin{equation}
    \label{flowGeneration}
    \featVar = \flowFunc_\flowParam(\baseVar) \text{~~where~~} \baseVar \sim p_{\baseParam}(\baseVar)\,
\end{equation}

where $\baseVar \in \mb{R}^\featDim$ and $p_{\baseParam}$ is commonly chosen as a uni-modal Gaussian.
By designing $T_{\flowParam}$ to be a \textit{diffeomorphism}, that is, a bijection where both $T_{\flowParam}$ and $T^{-1}_{\flowParam}$ are differentiable, We can compute the likelihood of the input $\featVar$ \textit{exactly} based on the change-of-variables formula~\cite{bogachev2007measure}: 
\begin{equation}
    \label{pdf_computation}
    p_{\flowParam, \baseParam}(\featVar) =  p_{\baseParam}(\flowFunc_{\flowParam}^{-1}(\featVar))|\det(J_{T_{\flowParam}^{-1}}(\featVar))|\,,
\end{equation}
where $J_{T_{\flowParam}^{-1}}(\featVar) \in \mb{R}^{\featDim \times \featDim}$ is the Jacobian of the inverse $\flowFunc_{\flowParam}^{-1}$ with respect to $\featVar$.
% The transformation $\flowFunc_\flowParam$ can constructed by composing a series of bijective maps denoted by $\subflowFunc_{\flowParam}^i$ 
%\flowFunc_\oodParam = \subflowFunc_1 \circ \subflowFunc_2 \circ ...  \circ \subflowFunc_\flowNum$.
When the target distribution is unknown but samples thereof are available, we can estimate the parameter $(\phi, \psi)$ by minimizing the forward \gls{kld}, which is equivalent to maximizing the expected \gls{ll}.% over the samples.

% \def\Jdet{|\det(J_{T_{\flowParam}^{-1}}(\featVar))|}
% \def\baseL{(p_{\baseParam}(\flowFunc_{\flowParam}^{-1}(\featVar)))}
% \begin{equation}
%     \label{mle}
%     \text{LL}(\flowParam, \baseParam) = \mathbb{E}_{p(\featVar)}\left[ \log \baseL + \log \Jdet \right]
% \end{equation}

% \paragraph{Topological Problems of \gls{nf}} % TODO: 
% \todo{explain this with the toy classification visualization}

\paragraph{Topological Mismatch}
However, since the base distribution $p_\baseParam(\baseVar)$ is usually a uni-modal Gaussian (e.g.\ \cref{fig:filaments:c}) and $\flowFunc_\flowParam$ is a diffeomorphism, problems arise for modeling data distribution with different topological properties. These include well-separated multi-modal distributions or distributions with disconnected components (e.g.,  \cref{fig:filaments:a}). For example, one can see that this leads to density filaments between the modes in \Cref{fig:filaments:b}. %, i.e. two arcs in this case. 
\citet{cornish2020relaxing} have shown that flows require a bijection with \textit{infinite bi-Lipshitz constant} when modeling a target distribution with disconnected support using a unimodal base distribution. Besides the diminishing modeling performance, this renders the bijection to be numerically "non-invertible", thus, causing optimization instability during training and unreliability of likelihood calculation~\cite{behrmann2021understanding}.

\subsection{Conditional Resampled Base Distributions}\label{sec:crsb}

%\paragraph{How to Solve It} 
One possible partial mitigation is by enriching the expressiveness of the flows. For example, by (a) increasing the number of layers or parameters, (b) using more complex base distributions, or (c) employing multiple \gls{nf}, e.g., \ mixtures of \gls{nf}. It is important to note that especially (a) and (c) may escalate the computational cost and memory burden. Moreover, scaling the normalizing flow's expressivity, (a) or (c), often does not increase the stability of the optimization%comes with less guarantee on stabilizing the optimization
~\cite{hagemann2021stabilizing} or 
%and securing 
the likelihood calculation. For these reasons, we pursue (b) and attempt to compensate for the complexity of the transformation with the elasticity of the base distribution. In other words, we use a more flexible but efficient base distribution to trade off a costly but sufficiently expressive bijection of the normalizing flow. This way we aim to capture desirable topological properties of the target distribution~\cite{jaini2020tails}.
% \begin{wrapfigure}[28]{r}{0.65\textwidth}
\begin{figure}[!ht]
    % \vspace{-20pt}
    % \glsunsetall
    % \hspace{12pt}
    \begin{tabular}{C{0.1625\textwidth} C{0.1625\textwidth} C{0.1625\textwidth} C{0.1625\textwidth} C{0.1625\textwidth}}
        Ground Truth: \(p(\featVar|y = 0)\) & Ground Truth: \(p(\featVar|y = 1)\) & Marginalized \gls{gm}: \(p_{\flowParam,\baseParam}(\featVar)\) & Marginalized \gls{crsb}: \(p_{\flowParam,\baseParam}(\featVar)\) & Margin.\ \gls{crsb} Base: \(p_{\baseParam}(\baseVar)\) \\
    \end{tabular}\\
    \centering
    \raisebox{0.35\height}{\rotatebox{90}{Two Moons}}
    \hspace{0.2pt}
    \includegraphics[width=0.1975\linewidth]{materials/toy/moons_gt_density_0.png}
    \hspace{-7pt}
    \includegraphics[width=0.1975\linewidth]{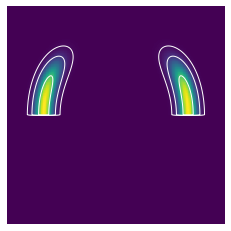}
    \hspace{-7pt}
    \includegraphics[width=0.1975\linewidth]{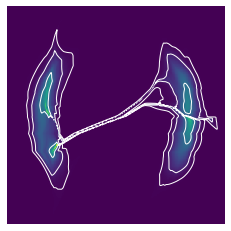}
    \hspace{-7pt}
    \includegraphics[width=0.1975\linewidth]{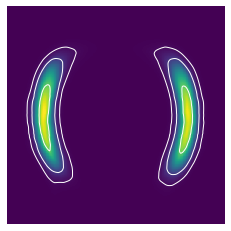}
    \hspace{-8pt}
    \includegraphics[width=0.2\linewidth]{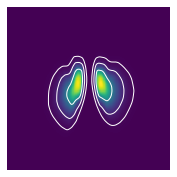}
    \\
    \vspace{-4pt}
    \raisebox{0.45\height}{\rotatebox{90}{Two Rings}}
    \includegraphics[width=0.1975\linewidth]{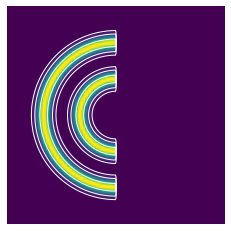}
    \hspace{-7pt}
    \includegraphics[width=0.1975\linewidth]{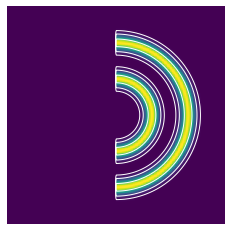}
    \hspace{-7pt}
    \includegraphics[width=0.1975\linewidth]{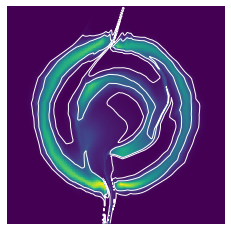}
    \hspace{-7pt}
    \includegraphics[width=0.1975\linewidth]{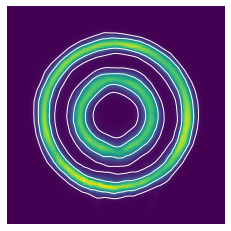}
    \hspace{-8pt}
    \includegraphics[width=0.2\linewidth]{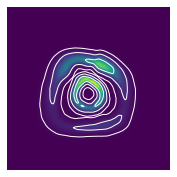}
    \\
    \vspace{-4pt}
    \raisebox{0.1\height}{\rotatebox{90}{Circle of Gauss.}}
    \hspace{0.2pt}
    \includegraphics[width=0.1975\linewidth]{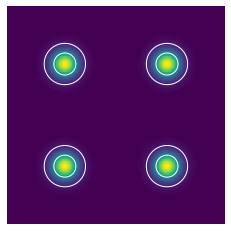}
    \hspace{-7pt}
    \includegraphics[width=0.1975\linewidth]{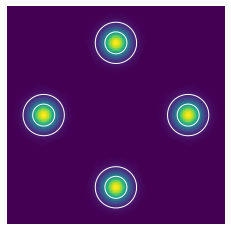}
    \hspace{-7pt}
    \includegraphics[width=0.1975\linewidth]{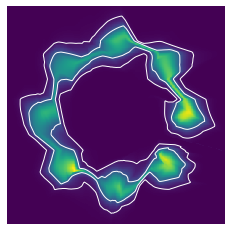}
    \hspace{-7pt}
    \includegraphics[width=0.1975\linewidth]{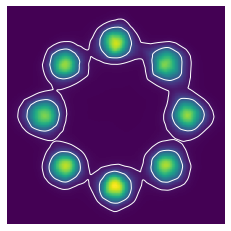}
    \hspace{-8pt}
    \includegraphics[width=0.2\linewidth]{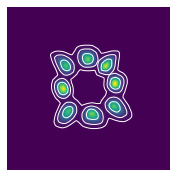}
    \caption{Visualization of density estimation %plotted in \([-3,3]^2\) 
    using Real NVP with class conditional \gls{gm}, where each class is modeled by a uni-modal Gaussian%, defined as \(\mathcal{N}(\boldsymbol{\mu}_c, \boldsymbol{\Sigma}_c)\)
    , and \gls{crsb} as well as the class-marginalized density for the base distribution of \gls{crsb}.  %We use 4 layers with \gls{crsb} and 5 layers with \gls{gm} to balance the compute of acceptance rate \(a_{\baseParam}\).
    \label{fig:density_toy}
    }
    % \glsresetall
\end{figure}
% \end{wrapfigure}
%
Following the prior work~\cite{fetaya2019understanding}, to model the fidelitous distribution of data with task-specific conditions, e.g. class labels, we use a class-conditional base distribution. 
This way we get similar benefits like combining multiple conditional flows (c), however, without having to burden the computational cost on marginalization over classes. \rebuttal{This is because, with (c), this operation requires repeated evaluation of the flows when each flow of the \gls{nf} mixture is class-conditional \cite{postels2021go}}. 
Even though a class-conditional distribution can specialize on a smaller fraction of the dataset containing similar instances, it will manifest in a multi-modal distribution. 
% To this end, we introduce the learnable class-conditional distribution, called \glsfirst{crsb}, in \Cref{sec:crsb}, which is capable of approximating complex target distributions and thus more accurately capturing their topological properties.

Therefore, we propose to capture the complex topological properties in the target distribution with a more expressive base distribution instead of the uni-model Gaussian. 
To the end, we introduce \gls{crsb} by extending a powerful unconditional base distribution \gls{rsb}~\cite{stimper2022resampling} with class-conditional modeling.
%\gls{rsb} replaces the uni-modal Gaussian with a learnable distribution that can approximate a complex distribution based on \gls{lars}~\cite{bauer2019resampled}. 
\gls{rsb} deforms a uni-modal Gaussian in a learnable manner to obtain more complex distributions via \gls{lars}~\cite{bauer2019resampled}. 
\gls{lars} iteratively re-weighs samples drawn from a proposal distribution $\pi(\baseVar)$, e.g. a standard Gaussian, through a learned acceptance function $a_{\baseParam}: \mathcal{R}^{\featDim} \rightarrow [0,1]$.
To reduce the computation cost in practice, this process is truncated by accepting the $T$-th samples if the previous $T-1$ samples get rejected. 
% The resulting distribution is as follows:
% \begin{equation}
%     \label{rsb_base}
%     p_{\baseParam}(\baseVar) = (1 - \alpha_T)\frac{a_{\baseParam}(\baseVar)\pi(\baseVar)}{Z} + \alpha_T\pi(\baseVar)
% \end{equation}, 
To take into account class-conditional information, we conditionalize the learnable acceptance function $a_{\baseParam}(\baseVar|y)$. 
As a result, we have the conditional base distribution: %by modifying the output dimension of the acceptance function  $a_{\baseParam}(\baseVar)$ to predict the acceptance rate for each class. 
%With this extension, we have the following conditional base distribution:
\begin{equation}
    \label{rsb_base}
    p_{\baseParam}(\baseVar|y) = (1 - \alpha_T)\frac{a_{\baseParam}(\baseVar|y)\pi(\baseVar)}{Z_y} + \alpha_T\pi(\baseVar),
\end{equation} 
where $a_{\baseParam}: \mathcal{R}^{\featDim} \rightarrow [0,1]^{C}$ and $\alpha_T = (1-Z_{y})^{T-1}$, where $Z_{y} \in \mathcal{R}$ is the normalization factor for $a_{\baseParam}(\baseVar|y)\pi(\baseVar)$. 
This factor can be estimated via Monte Carlo Sampling.

In \Cref{fig:density_toy}, we contrast the density estimation capabilities of \gls{nf} with the common \gls{gm}~\citep{li2022out,fetaya2019understanding} base distribution and our \gls{crsb} on three tasks with class-conditional structure using an appropriate learning objective (see next section). We find that our \gls{crsb} learns appropriate topology-matching base distributions (right outer column) and as a result, the respective \gls{nf} do not have adverse effects like filaments between the modes. 

% \paragraph{Discussion} There is another way to conditionalize the acceptance function by making the input $\baseVar$ conditioning on the class.  
% However, this way requires higher computation overhead, i.e. multiple forward passes for each class when we need a marginalized base density for \gls{ood} detection during inference.
% On the other hand, inspired by the \gls{gm}, we also tested class conditional Gaussians for the proposal distribution $\pi(\baseVar)$.
% However, the hypothesized expressiveness did not transfer to performance increase in preliminary experiments, we leave the investigation of this as future work. 

% \subsubsection{\gls{ib} for conditional \gls{nf}}
% Class conditional information is valuable for accurate density estimation of data with semantic class labels e.g. $p(\featVar|\bb{y})$.
\subsection{Training with Information Bottleneck}\label{sec:ib}
Unfortunately, directly training \gls{nf} with a conditional base distribution can lead to underperformance as observed in experiments (see \cref{tab:ood_od} and appendix) and reported by~\citet{fetaya2019understanding}.
We attribute this to the lack of explicit control for the balance between generative and discriminative modeling in the likelihood-based training objective of \gls{nf}.
To alleviate this, we train the normalizing flow with a class-conditional base distribution using the \gls{ib} objective~\cite{tishby2000information}. 
To abuse the notations, we denote random variables by capital letters such as $\featVarCap$, $\baseVarCap$, $Y$, and their realizations by lowercase letters such as $\featVar$, $\baseVar$, $y$. 
The \gls{ib} minimizes the \gls{mi} $I(\featVarCap, \baseVarCap)$ between $\featVarCap$ and $\baseVarCap$, while simultaneously maximizing the \gls{mi} $I(\baseVarCap, Y)$ between 
 $\baseVarCap$ and $Y$. 
Intuitively, the \gls{ib} trades off between the objectives of modeling the class conditional information $p(\featVar|y)$ with the marginalized density $p(\featVar)$, thus allowing to leverage the class-conditional structure to facilitate more effective density estimation for data characterized with semantic classes.
% Intuitively, this allows for learning a specialized and minimal class-conditional base distribution that may deviate from the class-conditional objective if it improves the marginalized distribution \(p(\featVar)\).

%\gls{ib}~\cite{tishby2000information} provides a principled and explicit way to coordinate the influences from two different variables, e.g. $\featVarCap$ and $Y$ for learning the latent variable $\baseVarCap$. \gls{ib} is an information theoretic approach for extracting relevant information measured by \gls{mi} about one variable $Y$ from another one $\featVarCap$. This relevant information is squeezed through a 'bottleneck' of the information flow between $\featVarCap$ and $Y$ in form of another latent variable $\baseVarCap$. Therefore, \gls{ib} strives to minimize the \gls{mi} $I(\featVarCap, \baseVarCap)$ between $\featVarCap$ and $\baseVarCap$ and simultaneously maximize the \gls{mi} $I(\baseVarCap, Y)$ between $\baseVarCap$ and $Y$ with a trade-off parameter to balance the two aspects.
% \begin{equation}
%     \label{ib}
%     \mathcal{L}_{IB} = I(\featVarCap, \baseVarCap) - \beta I(\baseVarCap, Y) 
% \end{equation}.

However, the \gls{ib} is not directly applicable to latent class-conditional distributions in \gls{nf} since the bijection \(\flowFunc_{\flowParam}\) is lossless by design. Thus, for trading off the class-conditional information with density estimation capabilities, we adapt the approach proposed by~\citet{ardizzone2020training} for our \gls{crsb}. Specifically, we inject a small amount of noise $\epsilon$ into the input $\featVarCap$ and hence $\baseVarCap_{\epsilon}= T^{-1}_{\flowParam}(\featVarCap + \epsilon)$. Further we define an asymptotically exact version of \gls{mi}, namely the \gls{ci} (more details in appendix):
%However, \gls{ib} is not directly ready for adoption as it is problematic to estimate the \gls{mi} between the input variable $\featVarCap$ and the latent $\baseVarCap$ of a flow due to the deterministic and bijective mapping $\flowFunc_\flowParam$, which raises the issue of invalidity of the definition of joint distribution between $\featVarCap$ and $\baseVarCap$.
% To this end, we adopt a technique proposed by~\citet{ardizzone2020training} which injects a small amount of noise $\epsilon$ into the input $\featVarCap$ for valid density definition and introduces an asymptotically exact version of \gls{ib} by defining another function $\hat{I}(\cdot, \cdot)$ to measure the relevant information between two random variables (more details in appendix):
\def\IBJdet{|\det(J_{T_{\flowParam}^{-1}}(\featVar+\epsilon))|}
\def\IBbaseL{(p_{\baseParam}(\flowFunc_{\flowParam}^{-1}(\featVar+\epsilon)))}
\def\condIBbaseL{(p_{\baseParam}(\flowFunc_{\flowParam}^{-1}(\featVar+\epsilon)|y))}
\def\SumCondIBbaseL{\sum_{y\prime}(p_{\baseParam}(\flowFunc_{\flowParam}^{-1}(\featVar+\epsilon)|y\prime))}
\begin{equation}
    \label{ib_nf}
    \mathcal{L}_{\text{IBNF}} =  CI(\featVarCap, \baseVarCap_{\epsilon}) - \beta CI (\baseVarCap_{\epsilon}, Y)  % \text{~~~~with}
\end{equation}
\begin{equation}
    \label{ci_xz}
    CI(\featVarCap, \baseVarCap_{\epsilon})=\mathbb{E}_{p(\featVar), p(\epsilon)}\left[ - \log {\sum_{y\prime}p_{\baseParam}(\baseVar_{\epsilon}|y\prime)} - \log \IBJdet \right],
\end{equation}
\begin{equation}
    \label{ci_yz}
     CI(\baseVarCap_{\epsilon}, Y) = \mathbb{E}_{p(y)}\left[ \log \frac{p_{\baseParam}(\baseVar_{\epsilon}|y)p(y)}{\sum_{y\prime}p_{\baseParam}(\baseVar_{\epsilon}|y\prime)p(y\prime)} \right],  % \text{~~~~and} 
\end{equation}
% \begin{equation}
%     \label{ci_xz}
%     CI(\featVarCap, \baseVarCap_{\epsilon})=\mathbb{E}_{p(\featVar), p(\epsilon)}\left[ - \log {(p_{\baseParam, \flowParam}(\featVar + \epsilon))} \right],
% \end{equation}
%
%Moreover, $\beta$ balances class conditional information with density estimation capabilities.
\begin{wrapfigure}[11]{r}{0.425\textwidth}
    \vspace{-10pt}
    % \hspace{8pt}
    \begin{tabular}{C{0.03\linewidth} C{0.3\linewidth} C{0.3\linewidth}}
     & \(a_{\baseParam}(\baseVar|y=0)\) & \(a_{\baseParam}(\baseVar|y=1)\) \\
    \end{tabular}\\\\
    \centering
    \vspace{-6pt}
    \raisebox{0.8\height}{\rotatebox{90}{w/o \gls{ib}}}
    \includegraphics[width=0.4\linewidth]{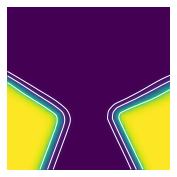}
    \hspace{-9.8pt}
    \includegraphics[width=0.4\linewidth]{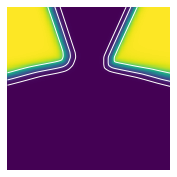}
    \\
    %\vspace{-3pt}
    \hspace{0.3pt}
    \raisebox{1\height}{\rotatebox{90}{w/ \gls{ib}}}
    \includegraphics[width=0.4\linewidth]{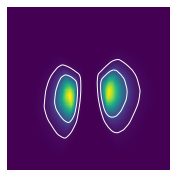}
    \hspace{-10.2pt}
    \includegraphics[width=0.4\linewidth]{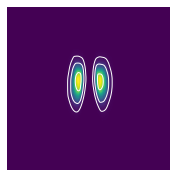}
    \caption{\gls{crsb} acceptance rate \(a_{\baseParam}(\baseVar)\) w/o and w/ \gls{ib} training for Two Moons.}
    \label{fig:rings_crsb_acceptance}
\end{wrapfigure}
where \(\baseVar_{\epsilon} = T^{-1}_{\flowParam}(\featVar + \epsilon)\), $p(\epsilon) = \mathcal{N}(0, \sigma^2 \mathcal{I}_{\featDim})$ is a zero-meaned Gaussian with variance $\sigma^2$, and $\beta$ trades off class information and generative density estimation.
With flexible conditional base distributions defined in Eq. \ref{rsb_base}, we can train the \textit{topology-matching} \gls{nf} with \gls{ib} by substituting \gls{crsb} into the conditional base probability $p_{\baseParam}(\baseVar|y)$ in Eq. \ref{ci_xz} and \ref{ci_yz}.
More noteworthy, we observed that the \gls{ib} is able to regularize the acceptance rate learning for \gls{crsb} to better assimilate the topological structure of the target distribution, leading to an overall improved performance on accurately approximating the complex target distribution (see \Cref{fig:rings_crsb_acceptance}). 

\subsection{Detecting OOD Objects}
\label{inference}
During test time, we detect the \gls{ood} data based on the predicted \glsfirst{ll}.
% For evaluation, to sidestep the influences of various selected thresholds, we measure the performance with \gls{auroc}. 
\rebuttal{To note that, only one forward pass is required to evaluate the acceptance function in \gls{crsb}. Practically, we use Monte Carlo sampling to estimate the normalization factor $Z$ offline so that no additional computation required for this during inference.}
We \textit{marginalize} the density over classes for the base distribution defined in Eq. \ref{rsb_base} and compute the final \gls{ll} given the logits $\pb{\featVar}$ from the test image: %and the predicted object bounding box $\{ \pb{x}$, $\pb{b} \}$:
\begin{equation}
    \label{test_ll}
    \operatorname{LL}_{test}(\pb{\featVar})=\log {\sum_{y\prime}(p_{\baseParam}(\flowFunc_{\flowParam}^{-1}(\pb{\featVar})|y\prime))} + \log |\det(J_{T_{\flowParam}^{-1}}(\pb{\featVar}))|.
\end{equation} We then expect \gls{ll} for \gls{id} objects to be higher than \gls{ood} ones.
%As we model the complicated \gls{id} with \gls{nf}, the expected \gls{ll} for \gls{id} objects should be higher than \gls{ood} ones. 
%A threshold to distinguish \gls{id} and \gls{ood} objects can be estimated from the training set. 
%===============================================================================
\section{Related Work}
\label{sec:related_work}

\textbf{Normalizing Flows} \gls{nf}~\cite{kobyzev2020normalizing} are a popular class of deep generative models. \gls{nf} have shown applicability in a variety of areas such as image generation~\cite{dinh2016density, kingma2018glow}, uncertainty estimation~\cite{charpentier2020posterior, charpentier2021natural, postels2020hidden} and \gls{ood} detection~\cite{kirichenko2020normalizing, nalisnick2018deep, lind2023out}. For \gls{nf}, one trend has been designing expressive flow-based architectures. Notable examples are affine coupling flows~\cite{dinh2016density, kingma2018glow}, auto-regressive flows~\cite{huang2018neural, durkan2019neural}, invertible ResNet blocks~\cite{chen2019residual} and ODEs-based maps~\cite{grathwohl2018ffjord}. The major focus of these works is on reducing computing requirements for Jacobian computations while ensuring that each mapping is invertible. 
Another research direction, currently emerging, is on addressing the topological mismatch~\cite{kobyzev2020normalizing, papamakarios2021normalizing} of \gls{nf}. Targeting this problem, some existing works attempt to increase the learning capacity of the transformation via mixture models~\cite{postels2021go}, latent variable models~\cite{cornish2020relaxing, dinh2019rad} or injecting carefully specified randomness~\cite{nielsen2020survae, wu2020stochastic}. 
These methods may be limited in their applicability to robotics because they either increase memory consumption by expanding the width of transformations or approximate the exact likelihood. Recently, these constraints have been addressed by improving the expressivity of the base distribution~\cite{stimper2022resampling, jaini2020tails}. In this paper, we build upon this class of methods since they only add slight computation overheads and thus are well suited for applications in robotics.

\textbf{Normalizing Flows for \gls{ood} Detection} \gls{nf} have been widely adapted for \gls{ood} detection due to its superior density estimation~\cite{yang2021generalized}. 
\rebuttal{For example, though with some counter-intuitive observations on raw data space~\cite{nalisnick2018deep}, \gls{nf} have demonstrated encouraging \gls{ood} detection results with additional refinements for raw data~\cite{ren2019likelihood, nalisnick2019detecting, jiang2022revisiting} or directly based on task-relevant feature embeddings~\cite{kirichenko2020normalizing, zhang2020hybrid, charpentier2023training, feng2023a}.
In this work, we directly apply \gls{nf} on the feature space.
To note that, another principle direction is to estimate the error bound for this task~\cite{chou2022safe}.}
Recently hybrid models~\cite{nalisnick2019hybrid,zhang2020hybrid, 9879060} have shown remarkable performance gain on \gls{ood} detection by modeling the joint distribution of both data and its class labels. Such works suggest that class labels can provide useful information. However, directly performing class conditional modeling with \gls{nf} for \gls{ood} detection results in performance degradation. \citet{tishby2000information, ardizzone2020training} mitigate such performance degradation by utilizing \gls{ib} for training \gls{nf}. This explicitly controls the trade-off between generative and discriminative modeling~\cite{mackowiak2021generative}.
However, these works on \gls{ood} detection utilize \gls{nf} without much concern for the fundamental topological problem as the first citizen. 
Therefore, complementary to these approaches, we examine the problem of topological mismatch of \gls{nf} for \gls{ood} detection.

\textbf{\gls{ood} Detection in Object Detectors} \gls{ood} detection research has focused on image classification~\cite{yang2021generalized}, which may be limited in relevance to robotic vision. 
In robotics, we may often need both categorization and localization of objects of interest. 
Therefore, we focus on object detection in open-set conditions here. 
In this domain, uncertainty estimation~\cite{gawlikowski2023survey} has been considered propitious for \gls{ood} detection but suffered from computation burdens on runtime~\cite{miller2018dropout,lee2020estimating,harakeh2020bayesod, feng2022introspective} or memory costs~\cite{lakshminarayanan2017simple}. 
To address this, instead of directly applying uncertainty estimation techniques for object detection~\cite{harakeh2020bayesod, lee2022trust}, another popular approach is to explicitly formulate the problem as \gls{ood} detection tasks~\cite{miller2021uncertainty, du2022siren, li2022out, kumar2023normalizing, du2022unknown}. Amongst them, \gls{nf} has been utilized as an expressive density estimator~\cite{li2022out, kumar2023normalizing}. However, despite the encouraging results, these approaches have not examined the problem of topological mismatch in \gls{nf}. As this might prevent additional performance improvements, this work examines the topology-matching \gls{nf} for \gls{ood} detection in object detectors.

\section{Experiments}
\label{sec:exp}
We next demonstrate the efficacy of our method. First, we evaluate 
%In this section, we first validate the proposed idea 
on synthetic density estimation for distributions with distinct topological properties.
We then evaluate the \gls{ood} detection performance on two object-detection data-sets adapted from their public counterparts~\cite{pascalVOC, lin2014microsoft} for open-set (OS) experiments: Pascal-VOC-OS and MS-COCO-OS based on Glow~\cite{kingma2018glow} and a pre-trained Faster-RCNN~\cite{ren2015faster} provided by ~\citet{miller2021uncertainty} for a fair comparison.
% For real robot deployment, we use the one-stage object detector Yolov7~\.
To showcase the practicality, we deploy the one-stage object detector Yolov7~\cite{wang2023yolov7} equipped with the proposed method on a real aerial manipulation robot along with the run-time and memory analysis.
\rebuttal{We empirically found that, to parameterize the acceptance function in \gls{lars}, a simple multi-layer perceptron (MLP) (2x128 for density estimation and 3x128 for object detection) is sufficient. We select the hyper-parameters (e.g., $T$, $\epsilon$, $\sigma$, $\beta$) based on the validation set.}  
More details can be found in the supplementary materials.
% Besides, we ablate the flow architectures to provide more evidence for the effectiveness of our method.

\paragraph{Datasets and Metrics}
For density estimation, there are three synthetic datasets: two moons, two rings, and a circle of Gaussians. 
We employ the \gls{kld} between the target and the model distributions to measure the performance.
For \gls{ood} detection, since existing object detection datasets are
not ready for fair evaluation~\cite{Dhamija_2020_WACV}, we strictly follow the experimental protocol in~\cite{miller2021uncertainty}. 
\rebuttal{For real robot deployment, we generate ~2$k$ synthetic images of two objects (a valve and a crawler robot) rendered based on their CAD models and additionally labeled ~2$k$ real images. 
~1$k$ synthetic images are used for training and another ~1$k$ for testing with all real images.} 
We use the \glsfirst{auroc} and the \gls{tpr} at different \gls{fpr} ($5\%$, $10\%$, $20\%$) as metrics for this task, as they represent the performance of the potential operating points for safety-critical applications, which requires the \gls{fpr} to be sufficiently low.

\subsection{Density Estimation}
 
\begin{wraptable}[8]{r}{0.5\textwidth}
%\begin{table}[!h]
\vspace{0pt}
\centering
% \ra{15}
\caption{Performance on density estimation for different flow architectures w.r.t.\ \gls{kld}, i.e., \(D_{\text{KL}}(p(\featVar, y) || p_{\flowParam, \baseParam}(\featVar, y))\). Better base distribution is highlighted in bold.\label{tab:toy_data}}
\begin{adjustbox}{width=\linewidth}
% \begin{tabular}{@{}r|rrr|rrrrr@{}}
\begin{tabular}{c|cc|cc}
% \toprule [1.25pt]
Flow architecture & \multicolumn{2}{c}{Real NVP} & \multicolumn{2}{c}{NSFs} \\[0.15cm]
% \cmidrule{3-5} \cmidrule{7-10} 
%\makecell[b]{Flow architecture \\ Base distribution} & \makecell[b]{Real NVP \\ GMM\_IB} &cRSB\_IB & \makecell[b]{NSFs \\ GMM\_IB} & cRSB\_IB \\
\makecell[b]{Base distribution} & MoG\_IB & cRSB\_IB & MoG\_IB & cRSB\_IB \\[0.1cm]
\midrule
Two Moons & 1.179 & \textbf{1.066} & 0.909 & \textbf{0.906}  \\ [0.1cm]
Two Rings & 2.032 & \textbf{1.704} & 1.647 & \textbf{1.602} \\ [0.1cm]
Circle of Gaussians & 2.335 & \textbf{1.667} & 1.766 & \textbf{1.653}  \\
% \bottomrule[1.25pt]
\end{tabular}
\end{adjustbox}
%\end{table}
\end{wraptable}
We compare the density estimation performance in \Cref{tab:toy_data} and provide qualitative results in \cref{fig:density_toy}.
We find that the \gls{crsb} base distribution consistently outperforms the class-conditional \glsfirst{gm}.
The performance improvement by \gls{crsb} can be generalized across two different \gls{nf} architectures, i.e.\ Real NVP and NSFs. %, validating that matching the topology of \gls{nf} can increase modeling performance on 2D density estimation.  

\subsection{OOD Detection in Object Detection}
\begin{wrapfigure}[10]{r}{0.5\textwidth}
   \centering
   \vspace{-38pt}
   \subfloat[\label{fig:voc_viz:a}]{\includegraphics[width=0.33\linewidth]{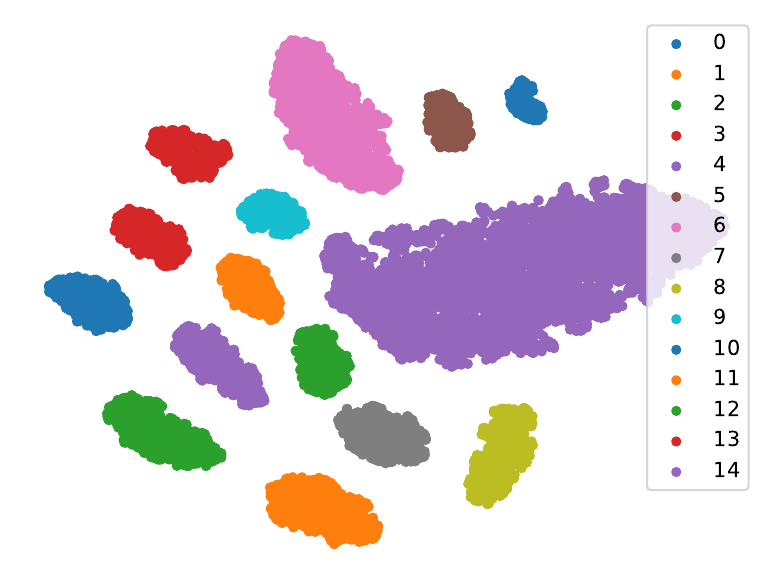}}
   \hfill
   \subfloat[\label{fig:voc_viz:b}]{\includegraphics[width=0.33\linewidth]{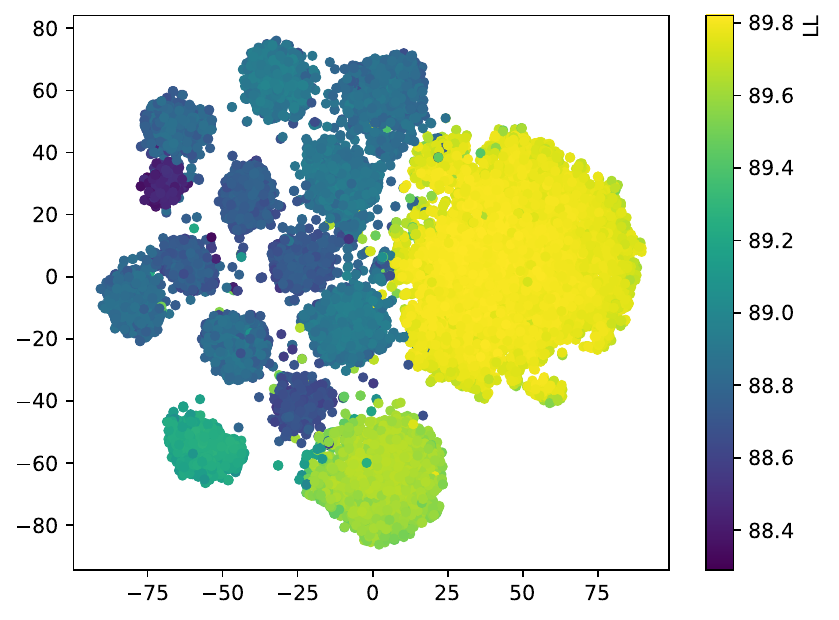}}
   \hfill
   \subfloat[ \label{fig:voc_viz:c}]{\includegraphics[width=0.33\linewidth]{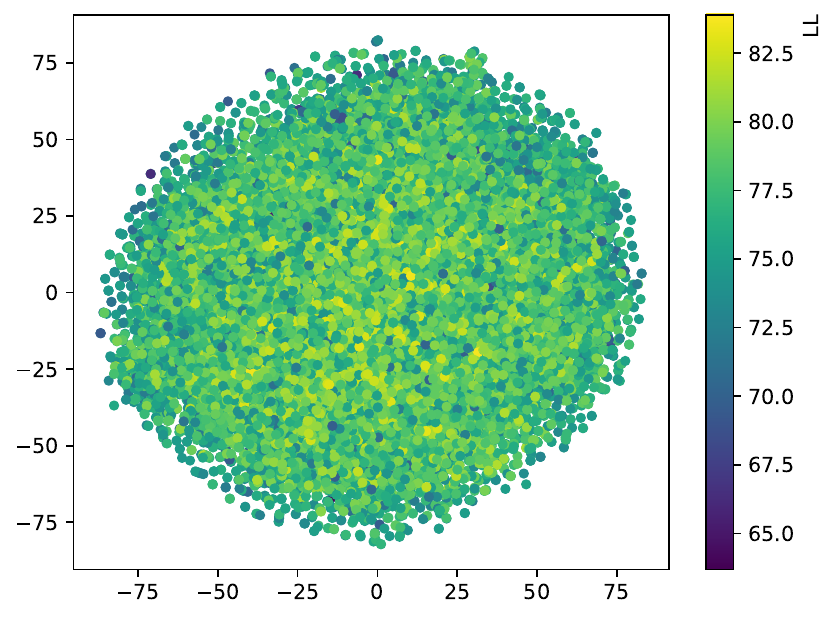}}
   \caption{t-SNE visualization for (a) feature embeddings from the object detector (b) latents of the proposed learned base distribution \gls{crsb} and (c) the uni-modal Gaussian on the training set of Pascal-VOC-OS.} %colored based on class labels.}
   \label{fig:voc_viz}
\end{wrapfigure}
%The main results of this experiment are presented in Table \ref{tab:ood_od}.
We compare our method (cRSB\_IB) with both flow-based and non-flow-based approaches. 
The latter consists of Mahalanobis Distance (MD) \cite{lee2018simple}, Relative Mahalanobis Distance (RMD)~\cite{ren2021simple}, GMMDet~\cite{miller2021uncertainty}, Softmax, Entropy and, their Deep Ensemble variants with five models~\cite{lakshminarayanan2017simple}.
%Here we compare the flow-based methods with GMMDet~\cite{miller2021uncertainty}, Softmax, Entropy and their Deep Ensemble versions~\cite{lakshminarayanan2017simple}.
Among flow-based approaches, we have six different base distributions, including unconditional ones (uni-modal Gaussian, \gls{gm}, \gls{rsb}) and their conditional variants (MoG\_CLS, cRSB\_CLS)~\cite{fetaya2019understanding} and \gls{gm} trained with \gls{ib} (MoG\_IB)~\cite{li2022out,ardizzone2020training}.
From Table \ref{tab:ood_od}, we can observe that flows with uni-modal Gaussian are able to provide satisfactory performance, i.e., better than most of non flow-based baselines, while flows with more expressive base distributions such as \gls{gm} and \gls{rsb} can bring more benefits on Pascal-VOC-OS than MS-COCO-OS. 
When trained with \gls{ib}, the more flexible conditional base distribution (cRSB\_IB) can mostly have greater performance gains (on both Pascal VOC and COCO) than its strong competitor (MoG\_IB) (only on COCO) in comparison with their counterparts without \gls{ib} (MoG\_CLS).
These results demonstrate the effectiveness of \gls{crsb} with \gls{ib} for \gls{ood} detection in complicated 2D object detection tasks.
We further provide the visualization from data before and after the flow transformation with different base distributions in \cref{fig:voc_viz}, evidencing the ability of matching complex topology of the target data distribution with \gls{crsb}.
% Though Ensemble GMMDet outperforms our method on TPR at 5\% and 10\%, it suffers from expensive computation and memory consumption, which is of disadvantage for applications in robotics.

\begin{table*}
\centering
% \vspace{-20pt}
\ra{1.32}
\caption{\gls{ood} detection performance on Pascal-VOC-OS and MS-COCO-OS datasets for different methods based on the Faster-RCNN from 3 random runs. The highest values are marked in \textbf{bold} and the second highest in \textit{italics}.}
\begin{center}
\label{tab:ood_od} 
\begin{adjustbox}{width=\textwidth}
\begin{tabular}{r|rrrrcrrrr}
% \toprule [1.25pt]
& \multicolumn{4}{c}{Pascal-VOC-OS} & \phantom{abc}& \multicolumn{4}{c}{MS-COCO-OS} \phantom{abc} \\
\cmidrule{2-5} \cmidrule{7-10} 
&AUROC &  & TPR at&  && AUROC & & TPR at & \\ 
& & $5\%$FPR & $10\%$FPR & $20\%$FPR &&  & $5\%$FPR & $10\%$FPR & $20\%$FPR \\ 
\midrule
Softmax & $0.901$ & $60.1$ & $72.8$ & $83.1$ && $0.882$ & $61.3$ & $70.6$ & $78.1$ \\
\rowcolor{gray!20}
Entropy & $0.905$ & $59.8$ & $72.9$ & $82.9$ && $0.903$ & $61.2$ & $70.6$ & $80.2$\\
MD \cite{lee2018simple} & $0.9$ & $54.1$ & $68.8$ & $83.3$ && $0.902$ & $57.2$ & $71.4$ & $85.5$ \\
\rowcolor{gray!20}
RMD \cite{ren2021simple} & $0.838$ & $15.2$ & $28.4$ & $77.4$ && $0.531$ & $1.7$ & $2.6$ & $7.1$\\
Ensemble Softmax~\cite{lakshminarayanan2017simple}& $0.885$ & $47.8$ & $72.6$ & $83.1$ && $0.898$ & $66.2$ & $73.5$ & $82.3$\\
\rowcolor{gray!20}
Ensemble Entropy~\cite{lakshminarayanan2017simple} & $0.887$ & $47.8$ & $72.5$ & $83.1$ && $0.906$ & $66.2$ & $73.5$ & $82.3$ \\
GMMDet~\cite{miller2021uncertainty} & $0.931$ & $70.7$ & $\textit{80.5}$& $\textit{89.3}$ && $0.924$ & $69.5$ & $80.2$ & $87.9$ \\
\rowcolor{gray!20}
% Ensemble GMMDet & $0.920$ & $\textit{73.6}$ & $79.6$ & $85.7$ && $0.927$ & $\textbf{80.7}$ & $\textbf{84.7}$ & $89.5$ \\
% \midrule[0.8pt]
Flows Gaussian & $0.915\pm0.002$ & $72.2\pm0.75$ & $77.8\pm0.89$ & $86.1\pm0.67$ && $0.924\pm0.001$ & $68.2\pm0.73$ & $81.2\pm0.61$ & $89.4\pm0.04$ \\
Flows MoG  & $0.919\pm0.002$ & $69.0\pm2.4$ & $77.0\pm2.5$ & $86.5\pm1.2$ && $0.925\pm0.001$ & $68.3\pm0.30$ & $80.5\pm0.50$ & $89.6\pm0.05$ \\
\rowcolor{gray!20}
Flows RSB~\cite{stimper2022resampling}& $0.924\pm0.003$ & $72.8\pm0.88$ & $79.3\pm1.0$ & $87.1\pm0.82$ && $0.925\pm0.001$ & $68.6\pm0.87$ & $81.3\pm0.31$ & $89.5\pm0.34$ \\
Flows MoG\_CLS~\cite{fetaya2019understanding} & $0.923\pm0.001$ & $69.2\pm1.5$ & $78.2\pm1.3$ & $88.5\pm0.82$  && $\textit{0.930}\pm0.001$ & $68.5\pm0.73$ & $\textit{82.2}\pm0.31$ & $\textit{89.7}\pm0.30$ \\
\rowcolor{gray!20}
Flows MoG\_IB~\cite{li2022out} & $\textit{0.934}\pm0.002$ & $73.1\pm1.3$ & $79.6\pm0.6$ & $87.8\pm0.2$ && $0.924\pm0.002$ & $\textit{71.1}\pm0.9$ & $79.6\pm0.46$ & $88.6\pm0.63$ \\
Flows cRSB\_CLS  & $0.919\pm0.001$ & $72.5\pm0.37$ & $78.8\pm0.27$ & $86.8\pm0.42$ && $0.924\pm0.001$ & $68.3\pm0.14$ & $81.1\pm0.30$ & $89.3\pm0.18$ \\
\rowcolor{gray!20}
Flows cRSB\_IB (ours)  & $\textbf{0.946}\pm0.003$ & $\textbf{78.5}\pm0.97$ & $\textbf{84.0}\pm0.83$ & $\textbf{90.8}\pm0.76$ &&  $\textbf{0.934}\pm0.002$ & $\textbf{73.3}\pm2.0$ & $\textbf{84.3}\pm0.40$ & $\textbf{91.3}\pm0.28$ \\
% \bottomrule[1.25pt]
\end{tabular}
\end{adjustbox}
\vspace{-20pt}
\end{center}
\end{table*}

\subsection{Real Robot Deployment}
\label{sec:real_boto}

\begin{figure}[!ht]
    \centering
    \vspace{-23pt}
    \subfloat[]
    {\includegraphics[width=0.345\linewidth]{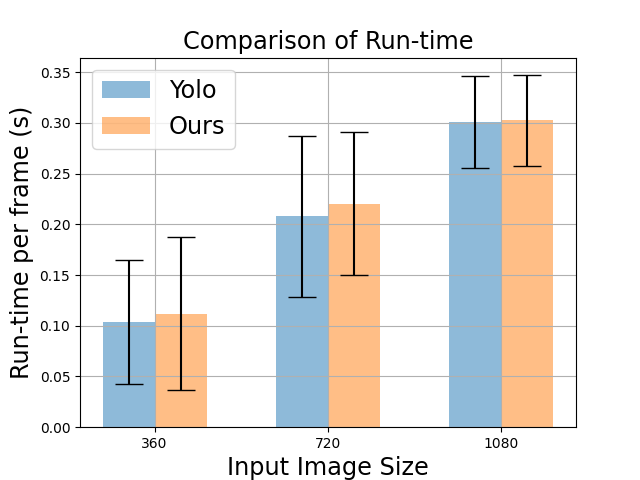}}
    \subfloat[]
    {\includegraphics[width=0.345\linewidth]{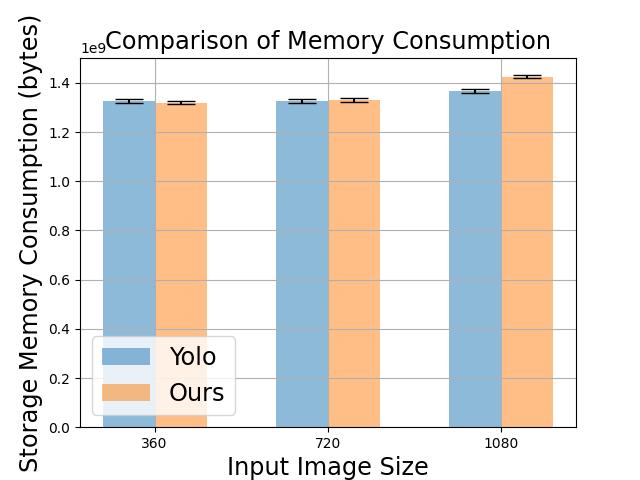}}
    \subfloat[\label{fig:real_robot_ood}]
    {\includegraphics[width=0.345\linewidth]{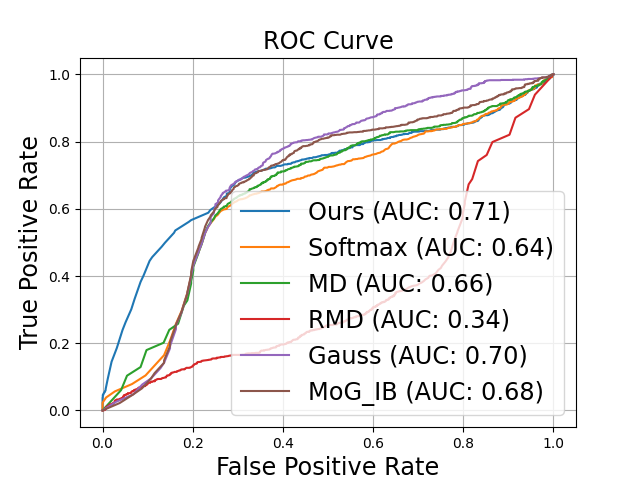}}
    \caption{Results from experiments on a real robot. Run-time, memory consumption, and ROC curve are reported. Compared to the vanilla Yolov7 , the proposed method does not yield significant computational costs, while providing performance gains in \gls{ood} detection.}
    \label{fig:real_robot}
\end{figure}

%Next, we validate the suitability of the proposed method for applications in robotics.
%To do so, we study our method by grounding it in an application scenario of robotic inspection and maintenance. 
% Within such field robotic applications, the \gls{ood} objects appear routinely due to the challenges of outdoor environments, which is crucial to avoid false positives. 
Next, we validate the applicability in an application of robotic inspection and maintenance, where it is crucial to avoid false positives of \gls{ood} objects that appear routinely in outdoor environments. 
\rebuttal{In this experiment, we train Yolov7 with only synthetic images of two objects (a valve and a crawler robot) and deploy on the robot around only real objects. The task is to identify the falsely detected real objects as \gls{ood} since they are from a distribution different to the synthetic ones}.
\rebuttal{Besides, the performance drop when compared with \cref{tab:ood_od} is potentially attributed to the "closer" OOD data because the synthetic images are rendered in a highly photorealistic manner}. 
However, our method still outperforms other baseline approaches in \cref{fig:real_robot_ood}, where ours can notably achieve higher \gls{tpr} around the low \gls{fpr} region, which are commonly used as operating points for the robot.
Computational efficiency is another important requirement. 
% The results are reported in Figure \ref{fig:real_robot}. 
We compare the runtime and space memory consumption against a vanilla Yolov7 using the NVIDIA's embedded GPU called Jetson Orin in \cref{fig:real_robot}. 
% Our approach is then applied to Yolov7 by training an additional \gls{ood} detector based on our topology-matching \gls{nf}. 
The results indicate that the computational overhead of having an \gls{ood} detector is relatively small when compared to the vanilla Yolov7. 
Overall, these experiments validate our claim that our method features efficient runtime inference and cost-effective memory consumption.

\section{Limitations}
\label{sec:conclusion}
The proposed method is envisioned to work on feature embeddings instead of raw data to counteract the \gls{nf} artifacts of assigning higher likelihoods to \gls{ood} data~\cite{papamakarios2021normalizing}. This leads to two limitations.
First, it's can't directly applied to the tasks/models that could not provide useful feature embeddings extracted from the raw data.
Second, its performance is restricted to the quality of features. As reported by previous work~\cite{miller2021uncertainty, li2022out}, learning more compact and centralized features can often lead to increased performance for \gls{ood} detection while feature collapse can be harmful to \gls{ood} detection.
%The major limitation for density-based \gls{ood} detection methods is the quality of the provided features used for density estimation. As reported by previous work~\cite{miller2021uncertainty, li2022out}, though modifications for existing models are inevitable, learning more compact and centralized features can often lead to increased performance for \gls{ood} detection while feature collapse can be disastrous for this kind of methods. Our proposed method, orthogonal to this direction, is compatible with more distinguishable features and promising to benefit from them.
Besides, there are two limitations during deployment. The first is the prolonged initialization time for calculating the normalization factor in \gls{lars} based on Monte Carlo sampling. This might not be friendly for applications that require instant response at the beginning. Moreover, the current version of the proposed method does not consider the sequential nature of observations at deployment.

% Another limiting factor of \gls{nf} is their computational cost. We mitigate the topological mismatch by enhancing the expressiveness of the base distribution. However, enhancing the base distribution also comes at a cost. Therefore, we seek to minimize the added computation overhead by balancing the cost of the bijection and base distribution. As a result, our method comes at negligible cost, when we equip the efficient Yolov7 object detector as shown in \cref{fig:real_robot}.
%Another limitation of the method is the resource trade-off between the transformation and the base distribution in \gls{nf}. The main idea of this work is to mitigate the topological mismatch by enhancing the expressiveness of the base distribution. However, investing too much resources such as large networks or computation-intensive inference on enriching the base distribution would cause again costly and resource-intensive burdens, which might be compensated by simply expanding the transformation. Therefore, we seek to minimize the added computation overhead in the proposed base distribution.

\section{Conclusion}
To endow robots with introspective awareness against \gls{ood} data, we propose the \gls{nf} equipped with effective yet lightweight \gls{crsb} and train with \gls{ib} objective. 
% With our learnable \gls{crsb} base distribution, \gls{nf} may bridge the gap between the topology of the base distribution and target distribution.
%With our learnable \gls{crsb} base distribution, 
Such \gls{nf} are able to mitigate the fundamental topological mismatch problem, facilitating more effective \gls{ood} detection capabilities.
We present empirical evidence that the proposed method achieves superior performance both quantitatively and qualitatively. 
To demonstrate the run-time efficiency and minimum memory overheads, we deployed on a real-robot system. Overall, we hope that the results of our work stemming from an enriched base distribution can push forward the direction of \gls{nf}-based \gls{ood} detection in robot learning.
%we focus on the density-based method with a proposed NF that can overcome its fundamental constraint by learning to match the topology in the base distribution to the target one, which is under-investigated in the literature.
%To this end, we introduce the conditional Resampled Base Distribution for \gls{nf} and train it with \gls{ib} for more effective \gls{ood} detection. We validated the proposed method on synthetic density estimation and public object detection benchmarks, achieving superior performance gain. To demonstrate the run-time efficiency and minimum memory overheads, we deployed on a real-robot system. Overall, the encouraging results and appealing practicality resulted from enriching base distribution shed light on further development for more effective and practical \gls{nf}-based \gls{ood} detection methods.

%===============================================================================

\clearpage
% The acknowledgments are automatically included only in the final and preprint versions of the paper.
\acknowledgments{We thank the anonymous reviewers for their thoughtful feedback. Jianxiang Feng and Simon Geisler are supported by the Munich School for Data Science (MUDS). Rudolph Triebel and Stephan
Gunnemann are members of MUDS.}

%===============================================================================

\bibliography{references}  % .bib

%===============================================================================
% \input{6_appendix}
%===============================================================================
\end{document}